\documentclass{article}

\PassOptionsToPackage{numbers, compress}{natbib}
\usepackage[final]{nips_2016}

\usepackage[utf8]{inputenc} % allow utf-8 input
\usepackage[T1]{fontenc}    % use 8-bit T1 fonts
\usepackage{hyperref}       % hyperlinks
\usepackage{url}            % simple URL typesetting
\usepackage{booktabs}       % professional-quality tables
\usepackage{amsfonts}       % blackboard math symbols
\usepackage{nicefrac}       % compact symbols for 1/2, etc.
\usepackage{microtype}      % microtypography
\usepackage{amssymb,amsmath,bm}
\usepackage{graphicx}
\usepackage{textcomp}
\usepackage{subcaption}
\providecommand{\norm}[1]{\left\|#1\right\|}

\title{Efficient Distributed Semi-Supervised Learning using Stochastic Regularization over Affinity Graphs}

\author{
Sunil Thulasidasan$^{1,2}$\\
\texttt{sunil@lanl.gov}
  \And
  Jeffrey Bilmes$^2$\\
\texttt{bilmes@uw.edu}\\
\And
Garrett Kenyon$^1$\\
\texttt{gkenyon@lanl.gov} \\
  \And
  \\
  $^1$Los Alamos National Laboratory \\
  $^2$Department of Electrical Engineering, University of Washington\\
}

\begin{document}
\maketitle

\begin{abstract}
  %We describe a graph-based semi-supervised learning framework in the context of deep neural networks that uses a graph-based entropic regularizer to favor smooth solutions over a graph induced by the data. %We consider graph embeddings constructed from  the input features as well as from dimensionality-reduced encodings obtained from the bottleneck layer of a deep neural network. 
 
We describe a computationally efficient, stochastic graph-regularization technique that can be utilized for  the semi-supervised training of deep neural networks  in a parallel or  distributed setting. We utilize a technique, first described in~\cite{sunil_mlslp_2016} for the construction of mini-batches for stochastic gradient descent (SGD)  based on synthesized partitions of an affinity graph that are consistent with the graph structure, but also preserve enough stochasticity for convergence of SGD to good local minima. We show how our technique allows a  graph-based semi-supervised loss function to be decomposed into a sum over objectives, facilitating data parallelism for scalable training of machine learning models.   Empirical results indicate that our method significantly improves classification accuracy compared to the fully-supervised case when the fraction of labeled data is low, and in the parallel case, achieves significant speed-up in terms of wall-clock time to convergence.  We show the results for both sequential and distributed-memory semi-supervised DNN training on a speech corpus. \end{abstract}

\section{Introduction}
\label{sec:intro}
Big data is often a deluge of unstructured, un-annotated and unlabeled data - a natural outcome of technological advances that have enabled data to be collected and disseminated with little effort and on large scales  but where annotating and labeling ground truth largely remains a time consuming, human effort. Semi-supervised learning (SSL) methods use  both labeled and unlabeled data to improve learning performance~\cite{chapelle2006semi}  and are especially useful in situations where labeled data is scarce
%Traditional fully supervised machine learning requires all training data to be labeled, which is usually done by humans, and thus expensive to obtain and difficult to scale. 
%In contrast, 
%Since unlabeled data can usually be  collected in a fully automated, scalable way, 
Such methods  leverage unlabeled data  by exploiting the similarity between labeled and unlabeled data by capturing this relationship via graphs, where the nodes represent both labeled and unlabeled points and the weights of the edges reflect the similarity between the nodes~\cite{zhu2005semi}. 

The main idea behind  graph-based SSL methods is that given a similarity metric, % simmilar poin; given a similarity metric (which does not depend on knowledge of the labels), 
the objective function constrains similar (i.e., nearby) nodes to have the same label by imposing a graph-neighbor regularization. This is effective because it forces the labels to be consistent with the graph structure, and the underlying manifold represented thereby.  The general form of the loss function in graph-based SSL has the following form
\begin{align}
\sum_{i=1}^L l(y_i,f(x_i)) + \lambda \sum_{i,j} \omega_{i,j} g(f(x_i),f(x_j)) \label{eq:gen_loss_fn}
\end{align}
where $f : \mathcal{X} \to \mathcal{Y}$ is the classifier mapping from input to output space. The first term in Equation~\ref{eq:gen_loss_fn}  the supervised loss function calculated on the labeled points, which can be a squared loss, hinge-loss or some measure of divergence between predictions and ground truth. The second term is the graph regularizer, where $\omega_{i,j}$ captures the similarity between points  $x_i$ and $x_j$.  $g(.)$  captures the discrepancy between output $f(x_i)$  and $f(x_j)$, incurring a  penalty when similar nodes have differing outputs. Additional regularizers such as the standard $\ell_2$ regularizer can also be applied to the above loss function to prevent overfitting.
Concretely, Let %$D^{\ell} =
$\{(\bf{x}_i,y_i)\}_{i=1}^{\ell}$ be the labeled training data
and %$D^{u} =
$\{\bf{x_i}\}_{i=\ell+1}^{\ell + u}$ be the unlabeled training data,
where $n = \ell + u$ so that we have $n$ points in total.  
%We denote
%by $\vartriangle_M$ the $M$-dimensional probability simplex (i.e., the
%set of all distributions over $M$ class labels). Let
%$\mathbf{p}_\theta (\mathbf{x}_i) \in \vartriangle_M$ represent the
%output vector of posterior probabilities dictated by $\theta$, the
%parameters of the classifier and $\mathbf t_i \in \vartriangle_M$ for
%$1 \leq i \leq \ell$ denote a probabilistic label vector for the
%$i$-th training sample.  
We  assume that the samples $\{ \mathbf{x}_i\}_i$ are
used to produce a weighted undirected graph $\mathcal{G} = (V,E, \bf W)$, where
$\omega_{i,j} \in \bf{W}$ is taken to be the similarity (edge weight)
between samples (vertices) $\mathbf x_i$ and $\mathbf x_j$.
We use the objective function defined 
in~\cite{subramanya2009entropic,malkin2009semi}, namely:
\begin{align}
J(\theta) &= \sum_{i=1}^l  \mathbf{D}( \mathbf{t}_i  \parallel \mathbf{p}_\theta (\mathbf{x}_i))  +
\gamma \sum_{i,j=1}^n \omega_{i,j} \mathbf{D}(\mathbf{p}_\theta (\mathbf{x}_i) \parallel \mathbf{p}_\theta (\mathbf{x}_j))   %\nonumber \\
 &+ \kappa \sum_{i=1}^n \mathbf{D}(\mathbf{p}_\theta (\mathbf{x}_i) \parallel
\mathbf{u})  + \lambda \norm{\theta},   \label{eq:loss_fn}
\end{align}
%where $\mathbf{u}$ %\in \vartriangle_M$
 %is the uniform distribution and 
  where $J(\theta)$  is the loss calculated over all samples. We use KL-divergence (denoted by $\mathbf{D}(.\parallel .)$) in our loss function since our output is a probability distribution over classes. The first term in the above equation is the supervised loss over the training samples, and the second term is the penalty imposed by the graph regularizer over neighboring pairs of nodes that favors smooth solutions over the graph.  The  third term  is an entropy regularizer ($\mathbf{u}$ is the uniform distribution) and favors higher entropy distributions to discourage degenerate solutions. The final term in Equation~\ref{eq:loss_fn} is the standard  $\ell_2$ regularizer to discourage overfitting. Note that the loss function, as such, is not directly decomposable as a sum over data points due to the presence of the graph regularizer and thus is not directly amenable to data parallelism. %in the context of a  multi-layered perceptron (MLP) with one hidden layer. 
This necessitates the implementation of data partitioning; such strategies for parallel machine learning in the fully supervised case  have been described, for example, in~\cite{smola2015,kaiwei2015nipsparallelworkshop}. %we describe algorithmic improvements for efficient and scalable graph regularization that can be applied to any parametric graph-based SSL framework. We
The presence of the graph regularizer term in the semi-supervised case necessitates a different approach, and in this work we study  the effectiveness of a stochastic  method  in the parallel setting, first described  in~\cite{sunil_mlslp_2016}. Our method constructs graph-based mini-batches by sampling the data using graph partitioning, but at the same time also preserves the statistical properties of the data distribution.   For the experiments in this work, we use  a deep neural network with a loss function given by the above equation but the method can be generalized to any parametric learner.

\subsection{Graph Partitioning for Objective Function Decomposition}
\begin{figure*}[ht]
	\centering
	%\hspace{-0.55in}
	\begin{subfigure}[b]{0.24\textwidth}
		\includegraphics[width=\textwidth]{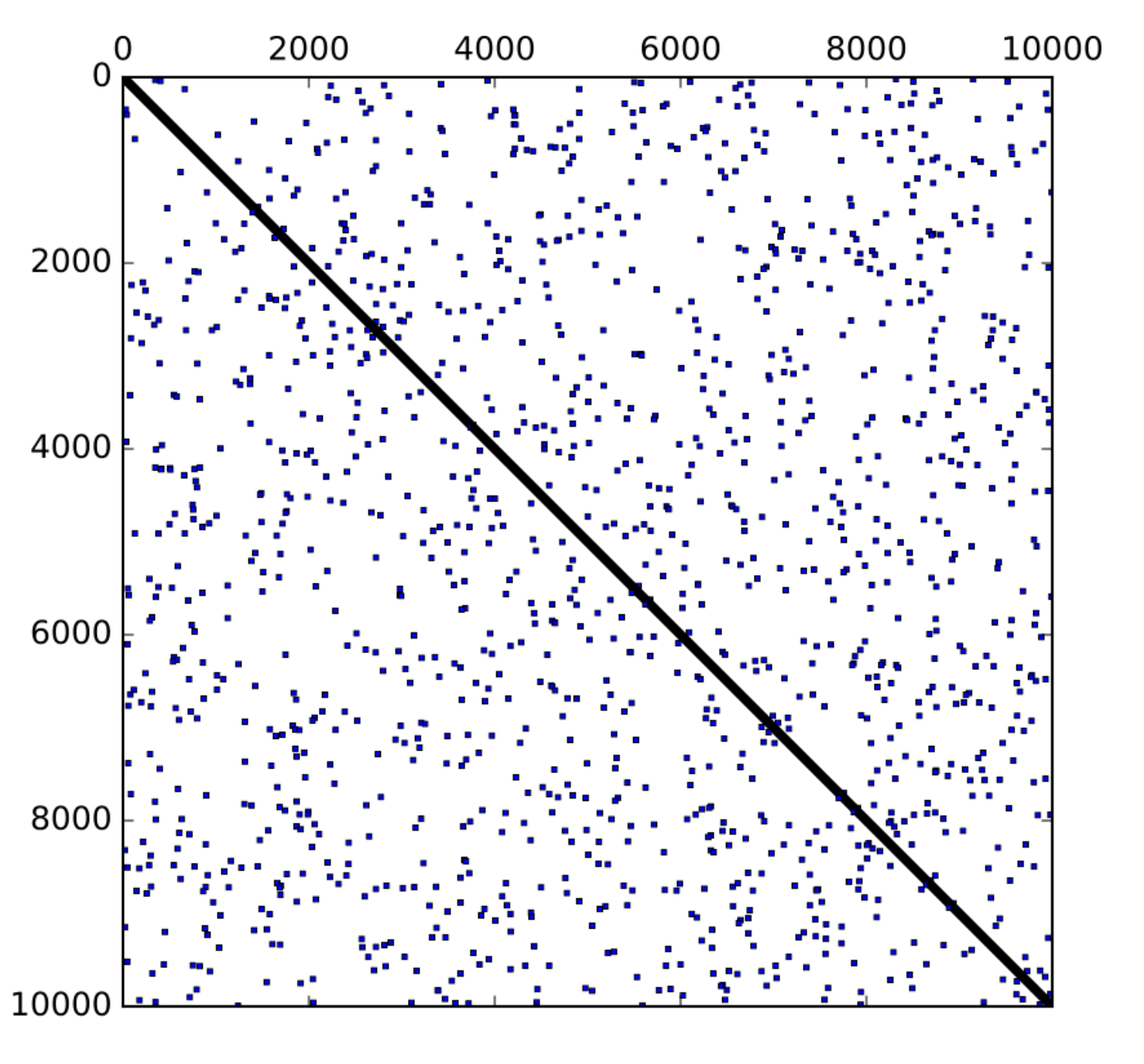}
		\caption{}
		\label{fig:mat_random}
	\end{subfigure}
	%quad
%\hspace{-0.45in}
	\begin{subfigure}[b]{0.24\textwidth}
		\includegraphics[width=\textwidth]{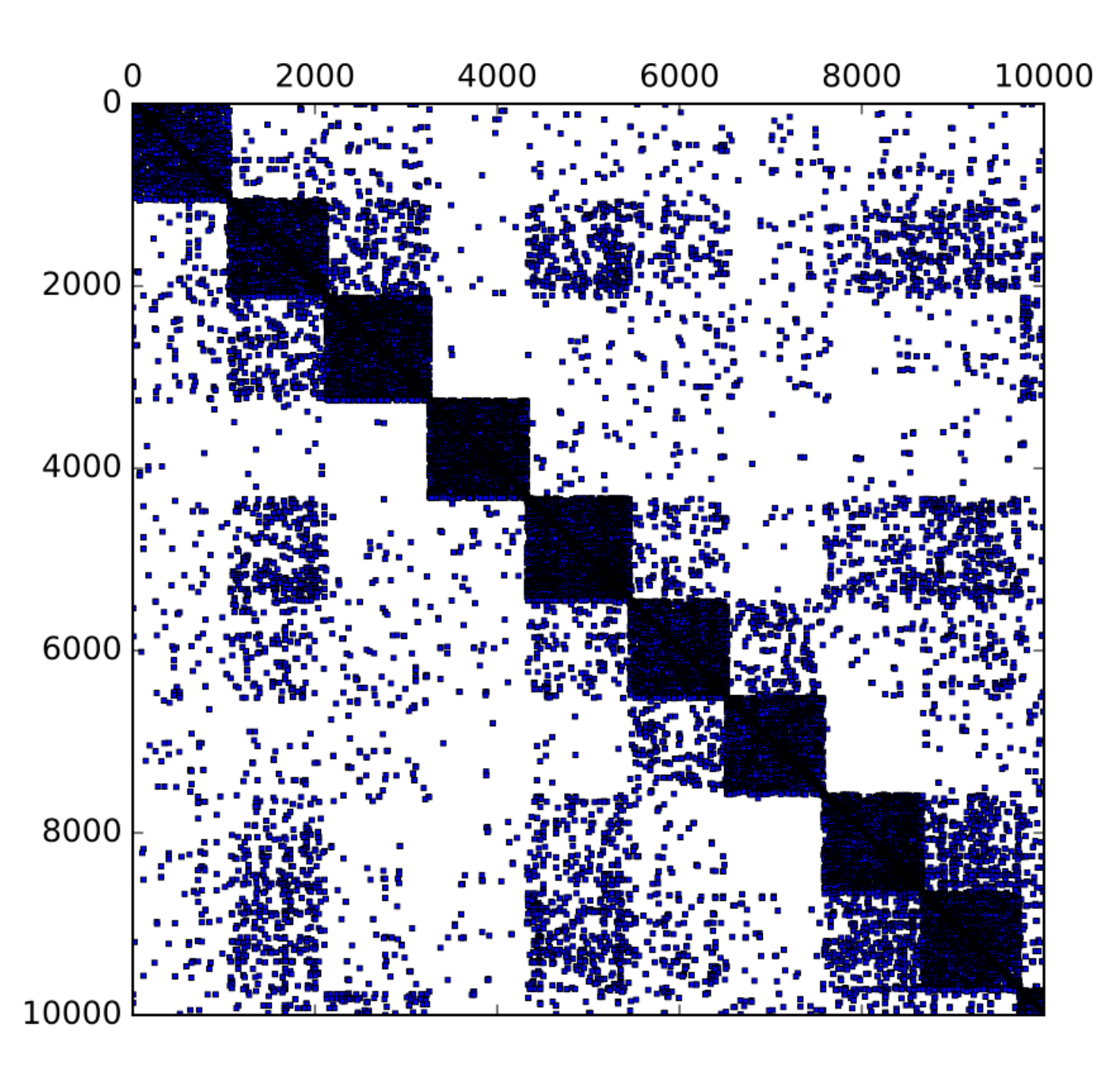}
		\caption{}
		\label{fig:mat_gpart}
	\end{subfigure}
	\begin{subfigure}[b]{0.32\textwidth}
		\includegraphics[width=\textwidth]{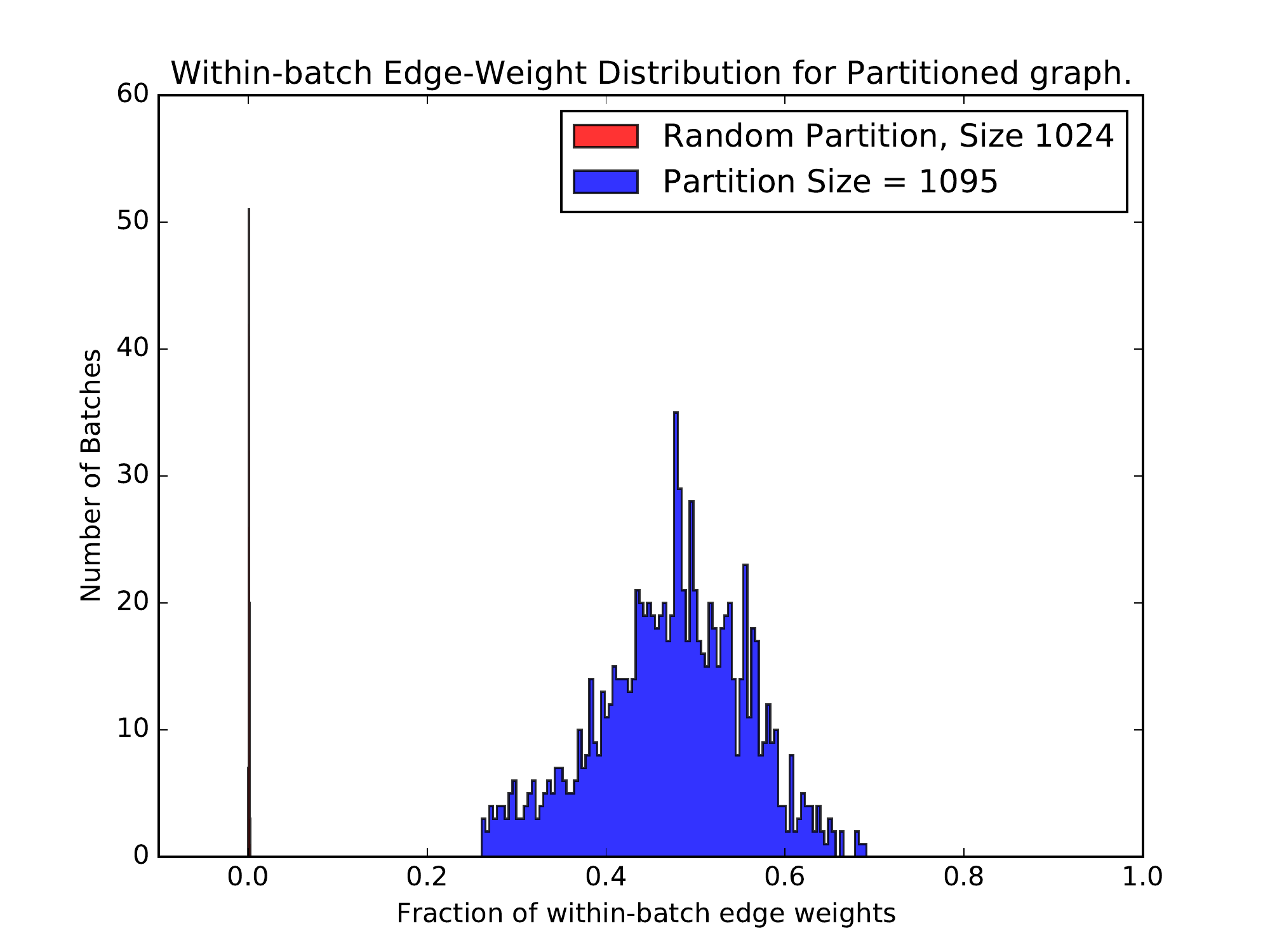}
		\caption{}
		\label{fig:bewr_dist_1}
	\end{subfigure}
%\hspace{-0.6in}
%\hspace{-0.5in}
\caption{(\ref{fig:mat_random})  $10,000\times10,000$ sub-block of the affinity matrix for randomly shuffled data. Choosing $1000\times1000$ blocks along the diagonal of such a  matrix would produce extremely sparse matrices.~(\ref{fig:mat_gpart}) Same affinity matrix re-permuted based on graph partitioning. While perform mini-batch computation we choose the diagonal blocks (dense squares).~(\ref{fig:bewr_dist_1}) Distribution showing the within-batch connectivity of graph-based batches (blue) vs randomly shuffled batches (spike). Nodes in randomly shuffled batches have almost none of the neighbors in the same batch}
\label{}
%\end{figure*}
\end{figure*}

Decomposing KL-divergence in Equation~\ref{eq:loss_fn} into entropy and cross-entropy terms and dropping the constant terms (w.r.t parameters), we can show that over one labeled point, the loss function becomes
 \begin{align}
%  J_i   &= D (\mathbf{t_i} \parallel \mathbf{p_i}) + \gamma \sum_{j=1}^n \omega_{ij}H^c(\mathbf{p}_i, \mathbf{p}_j) - (\kappa + \gamma \sum_{j=1}^n \omega_{ij}) H(\mathbf{p}_i) + \kappa \log K + \lambda \norm{\theta} 
J_i  = H^c(\mathbf{t}_i, \mathbf{p}_i) + \gamma \sum_{j=1}^n \omega_{ij}H^c(\mathbf{p}_i, \mathbf{p}_j) - (\kappa + \gamma \sum_{j=1}^n \omega_{ij}) H(\mathbf{p}_i)  + \lambda \norm{\theta}
\end{align}  

where $H$ and $H^c$ are, respectively, the entropy and cross-entropy. Since we are dealing with a non-convex objective function, and a moderately large data set ($\approx$ 1 million training samples), we use stochastic gradient descent  to optimize our objective function. We also use mini-batches to improve the gradient quality, and further, use larger mini-batches (size set either to 1024 or 2048) for better computational efficiency on GPUs.  In order to converge to good local minima, traditional SGD methods require randomly shuffling the data before constructing the mini-batches; this, however, poses a serious problem for our objective function. To see this,  consider the terms involving graph regularization from our decomposed objective function, calculated over each point:
 \begin{align}
%  J_i   &= D (\mathbf{t_i} \parallel \mathbf{p_i}) + \gamma \sum_{j=1}^n \omega_{ij}H^c(\mathbf{p}_i, \mathbf{p}_j) - (\kappa + \gamma \sum_{j=1}^n \omega_{ij}) H(\mathbf{p}_i) + \kappa \log K + \lambda \norm{\theta} 
G_i  =  \gamma \sum_{j=1}^n \omega_{ij}H^c(\mathbf{p}_i, \mathbf{p}_j) - \gamma \sum_{j=1}^n \omega_{ij} H(\mathbf{p}_i)  
\end{align}
For the graph regularization term to have any effect at all, the $w_{ij}$'s corresponding to the points in the mini-batch have to be non-zero. For a randomly shuffled data-set, given that the graph is very sparse (since each of the $\approx$ 1 million points only has a little more than 10 neighbors), the chunk of the affinity matrix corresponding to the mini-batch will be extremely sparse, implying that graph regularization will fail to take place on most computations.  One way to fix this is, for a given mini-batch, to loop over all the neighbors for each point in the mini-batch, but this prevents us from doing efficient matrix-matrix multiplications and completely degrades performance negating any benefits of using fast processors like GPUs.
Thus, for the graph regularizer to be effective in a computationally efficient way, our mini-batches need to reflect the structure of the graph. To do this, we partition our affinity graph into $k$ balanced parts (by minimizing edge-cut)
 %(i.e, given $k$,  we want to minimize the number of edges between partitions), and for this we use the METIS graph partitioning library~\cite{karypis1998multilevelk}. Optimal partitioning is NP-hard and METIS uses  heuristics based on recursive bisectioning to produce  $k$-way balanced partitions.  
which results in a re-permuted affinity matrix that has a dense block-diagonal structure as shown in Figure~\ref{fig:mat_gpart}; contrast this with the affinity matrix before partitioning in Figure~\ref{fig:mat_random} where most entries over a $1000\times1000$ block (corresponding to a mini-batch size of 1000) are zero.

%\begin{figure*}[ht]

%\begin{figure*}[ht]
%	\centering
%	\includegraphics[scale=0.5]{plots/bewr_dist.pdf}
%	\caption{Distribution showing the within-batch connectivity of graph-based batches (blue) vs randomly shuffled batches (spike). Nodes in randomly shuffled batches have almost none of the neighbors in the same batch}
%	\label{fig:bewr_dist_1}
%\end{figure*}
Dense mini-batches imply that most of the neighbors of the nodes within a mini-batch are contained within the same batch. More formally, let $\mathcal{N}_i$ represent the set of neighbors of node $i$ and $\mathcal{C}_i \subseteq \mathcal{N}_i$ be the set of neighbors of a node $i$ that are within the same batch. Let $\mathcal{M}_j$ be the set that represents mini-batch $j$.  We define the within-batch connectivity of $\mathcal{M}_j$ as
\begin{align}
 c_j = \frac{\sum_{i \in \mathcal{M}_j} |\mathcal{C}_i   |}{\sum_{i \in \mathcal{M}_j} |\mathcal{N}_i|}, j = {1,2,3 \dots k}  \label{eq:connectivity}
\end{align}
In the randomly shuffled (pre-partitioning scenario) we expect most of the $c_j$'s to be close to zero, while for graph partitioned mini-batches, we expect a relatively higher $c$. Figure~\ref{fig:bewr_dist_1} shows this distribution for the random minibatches (seen as a sharp spike near 0) and for graph-partitioned mini-batches (in blue). Partitioning gives us an efficient way of computing our objective function: given a matrix permutation induced by the graph partitioning, we re-permute the affinity and data matrices accordingly. Then, during each mini-batch computation, we calculate the objective function and gradients on these partitions. Note that graph-partitioning is a pre-processing operation, and only done once before training commences.

\section{Issues for Stochastic Optimization }
Theoretically, SGD, gives us an unbiased estimate of the true gradient, but only if the data is sampled  form the true distribution. Graph partitioned mini-batches violate this assumption, and cause high variance in gradients and prevent convergence to good solutions.  For a detailed discussion see~\cite{sunil_mlslp_2016}.

\subsection{Improving SGD Convergence using Graph-Synthesized Meta-batches}
We use the batch batch construction algorithm described in~\cite{sunil_mlslp_2016}  to construct {\em meta-batches} from smaller, homogeneous graph-partitionied mini-blocks. The resulting meta-batches are both diverse and preserve sufficient connectivity for graph regularization. The reader is referred to Sections 4.1 in~\cite{sunil_mlslp_2016} for details.

\begin{figure}
\centering
%	\begin{subfigure}[t]{0.32\textwidth}
%		\includegraphics[width=\textwidth]{plots/entropy_dist_0.pdf}
%		\caption{}
%		\label{fig:mb_entropy_dist}
%	\end{subfigure}
	\begin{subfigure}[t]{0.35\textwidth}
		\includegraphics[width=\textwidth]{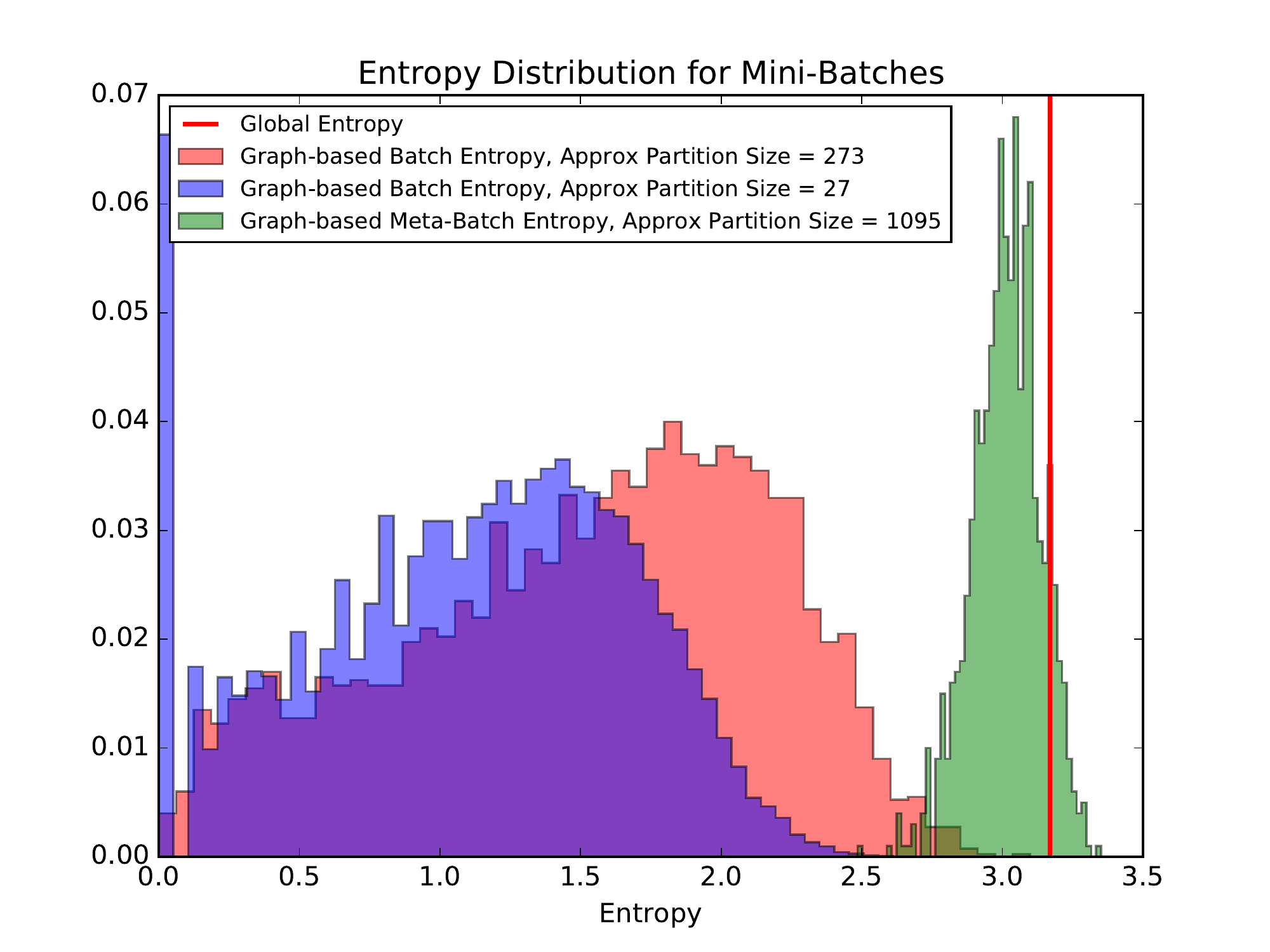}
		\caption{}
		\label{fig:mb_entropy_dist}
	\end{subfigure}
	\quad
	\begin{subfigure}[t]{0.35\textwidth}
		\includegraphics[width=\textwidth]{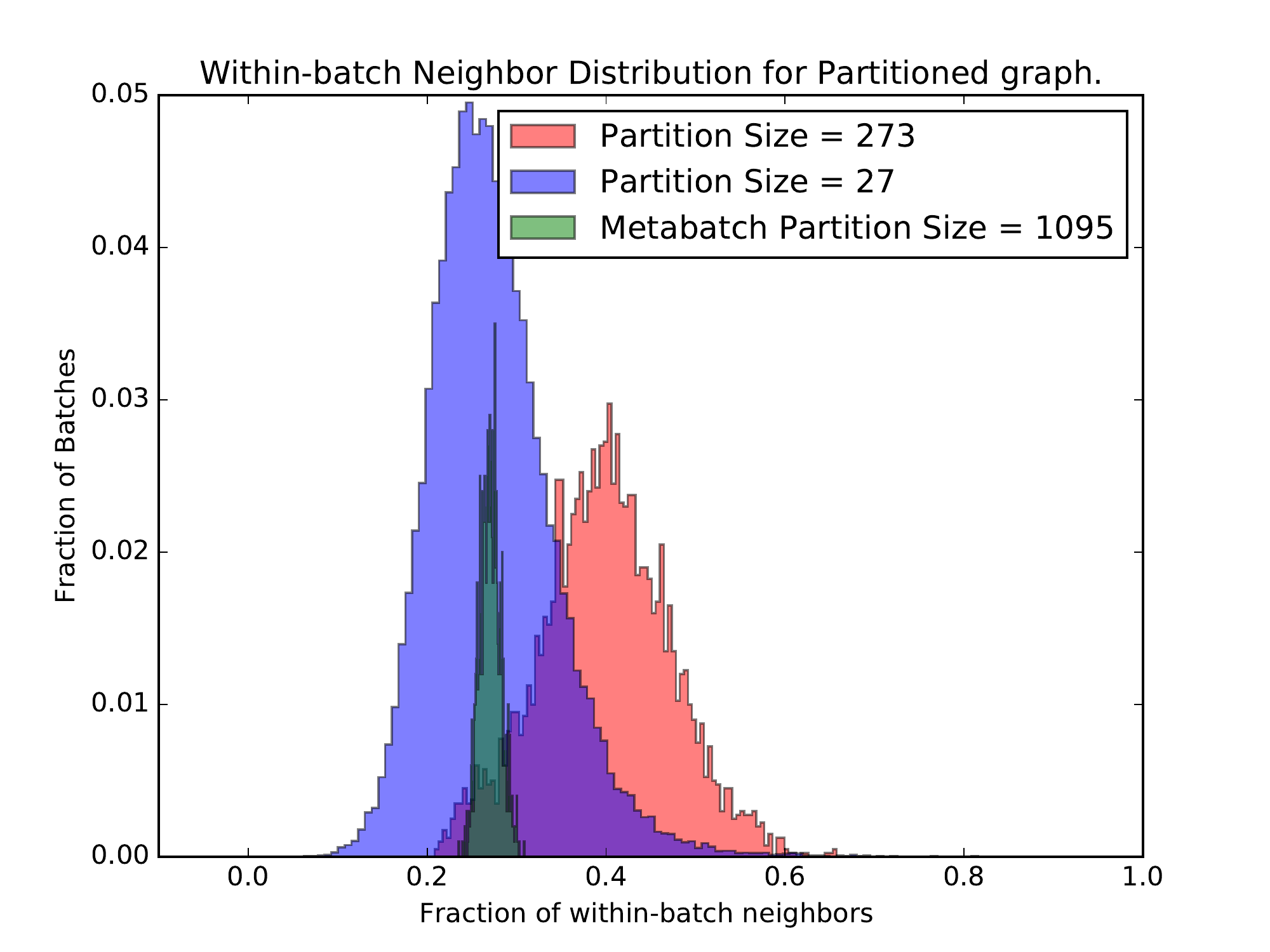}
		\caption{}
		\label{fig:mb_bnr_dist}
	\end{subfigure}
\caption{(\ref{fig:mb_entropy_dist}) Entropy distribution of mini-batches (calculated over labels in a mini-batch) for randomly shuffled mini-batches and graph-partitioned minibatches. Shuffled mini-batches have a tight distribution, very close to the entropy of the entire data set (shown as the vertical red line).~(\ref{fig:mb_entropy_dist}) Entropy distribution for meta-batches, compared to graph-partitioned batches. (~\ref{fig:mb_bnr_dist}) Metabatches preserve the average neighbor connectivity score of the smaller-partitions from which they are formed  }
\label{}
\end{figure}

The plots  shown in Figure~\ref{fig:mb_bnr_dist}  depicting histogram of connectivity and diversity (measured in terms of label entropy). The green histogram which shows distribution for the meta-batches has approximately the same mean as the mini-partitions (blue histogram) from which it is formed, but a much lower variance. %We also compare it to a larger batch of size 270 to show how batch sizes affect neighbor connectivity.
Overall meta-batches approximate the entropy of the global distribution while preserving sufficient connectivity for similarity regularization.

To enable regularization across meta-batches in a computationally efficient manner, at each iteration in an epoch, a meta-batch is also regularized with another randomly chosen meta-batch. See Section 4.2 in ~\cite{sunil_mlslp_2016} for details.

\subsection{Decomposing Loss Function for Parallel Training}

In Section~\ref{sec:intro} we noted how our original objective function was not easily decomposable as a sum over data points (or mini-batches) due to the graph regularization term. Using the techniques described above, we can facilitate data parallel training by considering again the objective function in terms of entropy and cross-entropy as given in Equation 3.
\begin{align}
%  J_i   &= D (\mathbf{t_i} \parallel \mathbf{p_i}) + \gamma \sum_{j=1}^n \omega_{ij}H^c(\mathbf{p}_i, \mathbf{p}_j) - (\kappa + \gamma \sum_{j=1}^n \omega_{ij}) H(\mathbf{p}_i) + \kappa \log K + \lambda \norm{\theta} 
J_i  = H^c(\mathbf{t}_i, \mathbf{p}_i) + \gamma \sum_{j=1}^n \omega_{ij}H^c(\mathbf{p}_i, \mathbf{p}_j) - (\kappa + \gamma \sum_{j=1}^n \omega_{ij}) H(\mathbf{p}_i)  + \lambda \norm{\theta}
\end{align}  

Now, for a given iteration over meta-batch $M_r$, and its randomly chosen neighbor meta-batch $M_s$, the points $i$ n the loss function are simply the labeled points in the concatenated batch $M_c = [M_r, M_s]$ while $j$ is the set of all points in $M_c$.  For a $k$-worker parallel training scenario, during each iteration,  there will be $k$ such meta-batches and  the gradients are calculated independently over these batches. Because  this is now (approximately\footnote{The decomposition is approximate since there will generally be cross-partition edges that we are ignoring in a given iteration}) decomposable as a sum over (concatenated) meta-batches, the technique can easily work within a parallel SGD framework. We present the results on a synchronized parallel SGD setup in the next section.

\section{Experiments}

\begin{figure*}
\centering
%	\begin{subfigure}{0.33\columnwidth}
%		\includegraphics[width=\columnwidth]{plots/gmlp_mb_bnr_vs_no_bnr.pdf}
%		\caption{Effect of meta-batches and stochastic graph regularization (SGR) on the performance of a graph-regularized MLP. Also shown are results of a base-line MLP, a fully-supervised learner that only uses labeled data.}
%		\label{fig:mlp_results}	
%	\end{subfigure}
%	\hfill
	\begin{subfigure}{0.32\columnwidth}
		\includegraphics[width=\columnwidth]{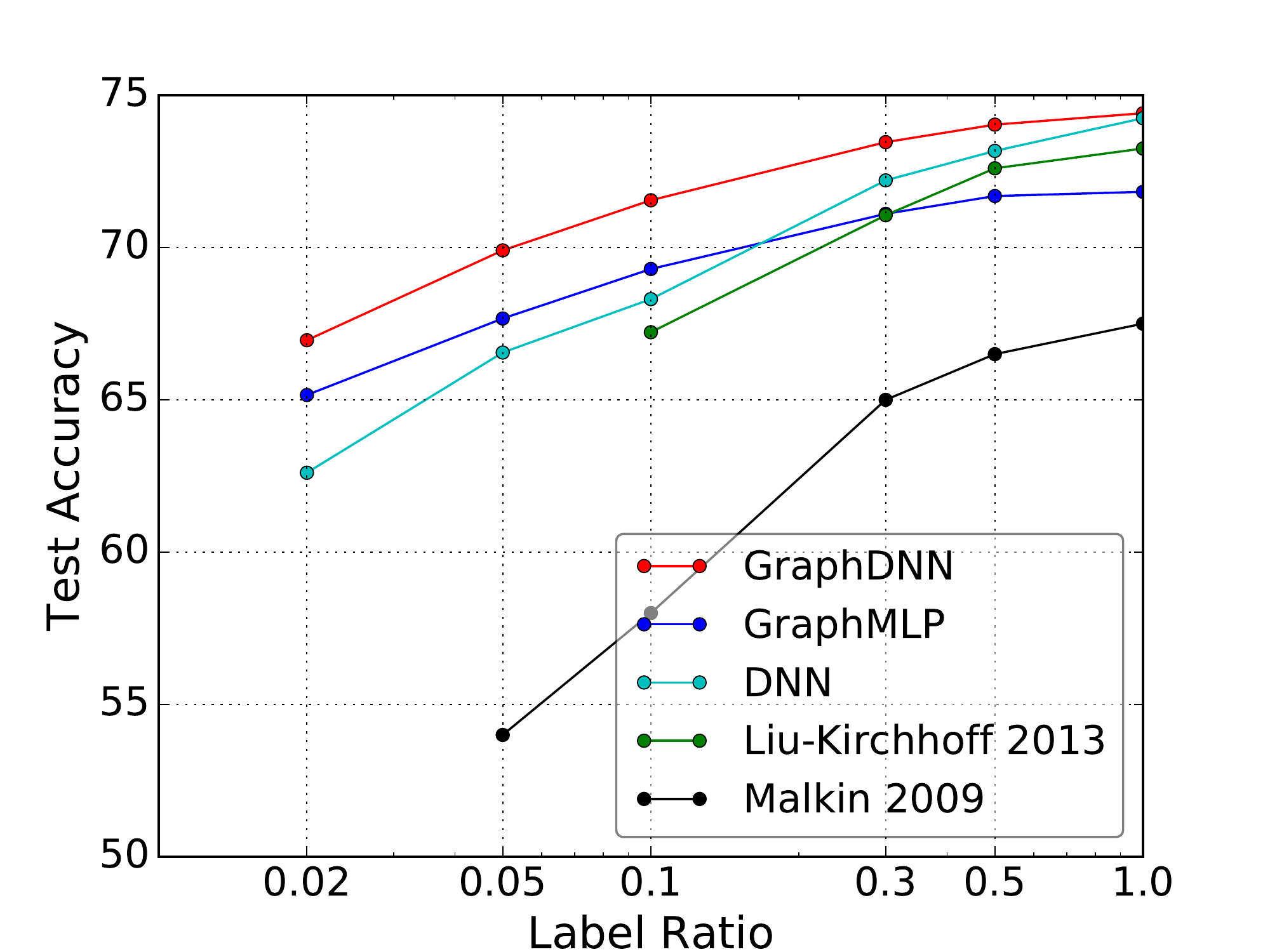}
%		\caption{Results for Graph Regularized DNN vs base-line DNN. The DNN used four hidden layers, each 2000 units, with dropout. Also shown are results for similar experiments in the literature that used the same TIMIT training and test data.}
		\caption{Sequential Case: Final test accuracy}
		\label{fig:dnn_results_2}
	\end{subfigure}
%	\hfill
%	\begin{subfigure}{0.33\columnwidth}
%		\includegraphics[width=\columnwidth]{plots/trade_off.pdf}
%		\caption{Trade-off between entropy and within-batch neighbor connectivity, and the performance (in terms of test error) at each of these scenarios. We used $5 \%$ of the labels, a batch-size of $\approx 1095$ and trained for  $100$ epochs. }
%		\label{fig:trade_off_results}
%		\end{subfigure}
%	\begin{subfigure}{0.4\columnwidth}
%		\includegraphics[width=\columnwidth]{plots/mxn_train_acc_plot.pdf}
%		\caption{Effect of meta-batches and stochastic graph regularization (SGR) on the performance of a graph-regularized MLP. Also shown are results of a base-line MLP, a fully-supervised learner that only uses labeled data.}
%		\label{fig:mxn_train_acc}	
%	\end{subfigure}
	\begin{subfigure}{0.32\columnwidth}
		\includegraphics[width=\columnwidth]{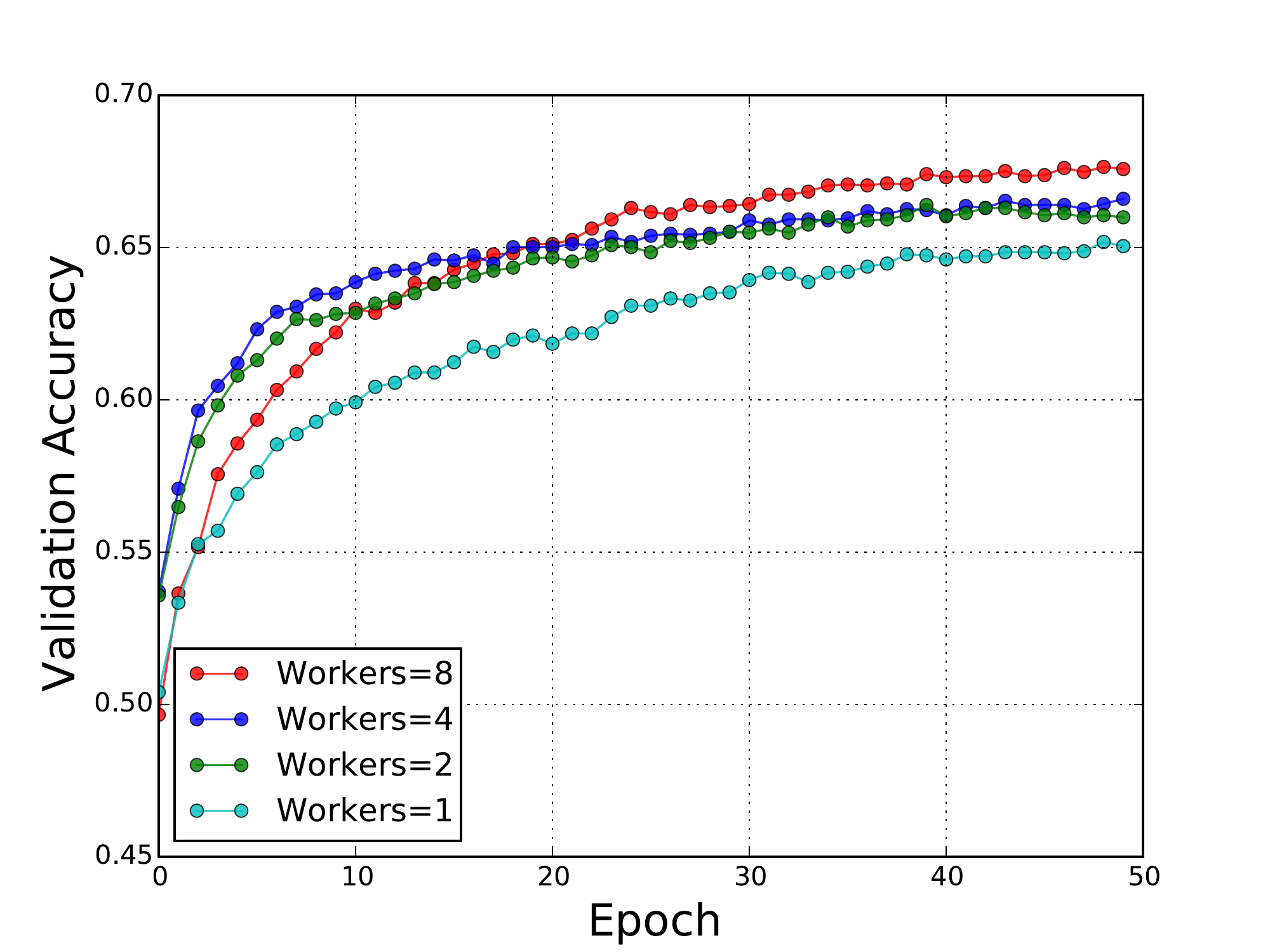}
%		\caption{Results for Graph Regularized DNN vs base-line DNN. The DNN used four hidden layers, each 2000 units, with dropout. Also shown are results for similar experiments in the literature that used the same TIMIT training and test data.}
		\caption{Parallel Case: Validation accuracy vs training epochs}
		\label{fig:mxn_val_acc}
	\end{subfigure}
	%\hfill
	\begin{subfigure}{0.32\columnwidth}
		\includegraphics[width=\columnwidth]{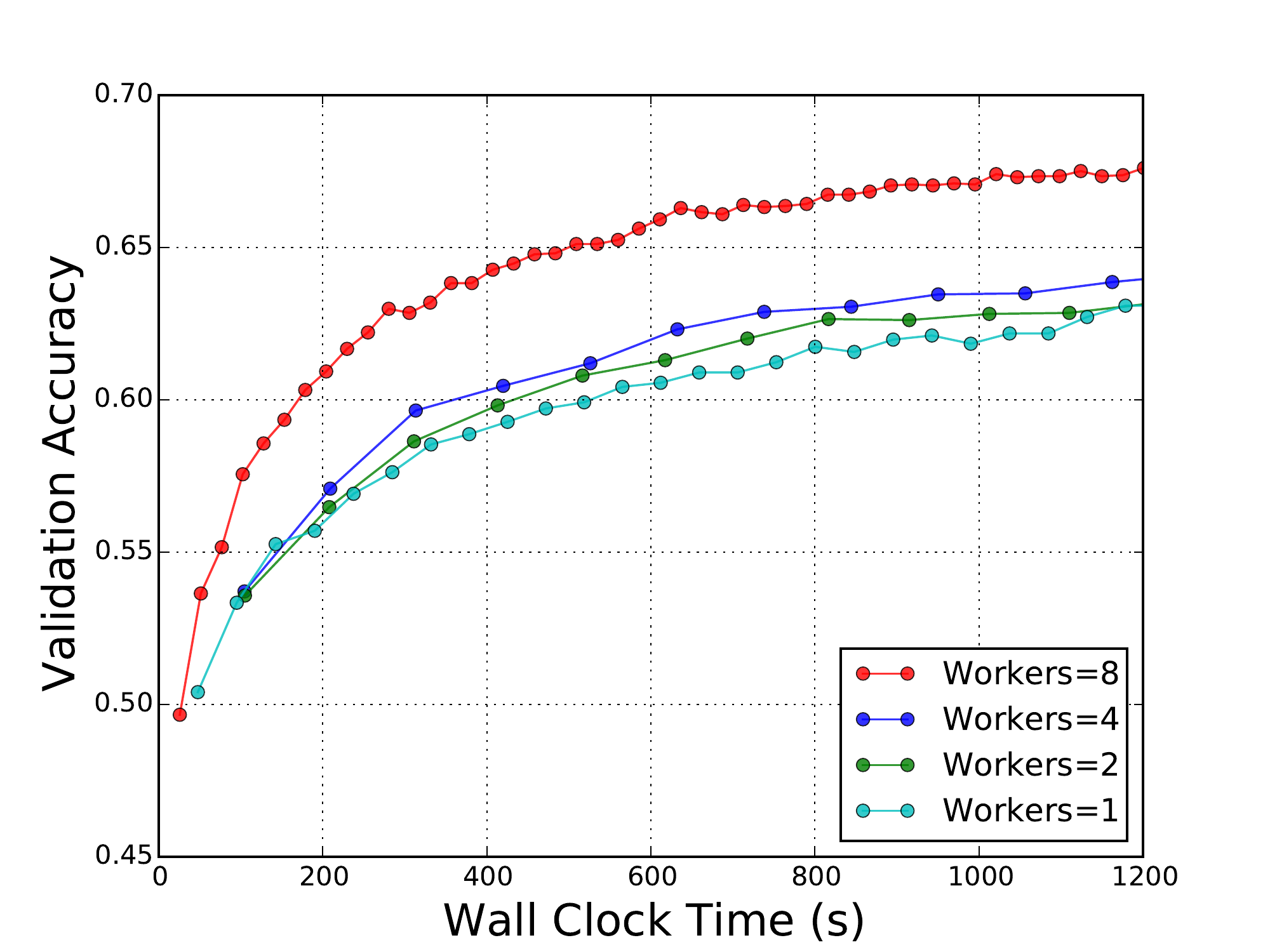}
%		\caption{Trade-off between entropy and within-batch neighbor connectivity, and the performance (in terms of test error) at each of these scenarios. We used $5 \%$ of the labels, a batch-size of $\approx 1095$ and trained for  $100$ epochs. }
		\caption{Parallel Case: Validation accuracy vs wallclock time}
		\label{fig:mxn_val_wall_clock}
		\end{subfigure}
\caption{Sequential and Parallel results for a Graph-Regularized DNN. Fig~\ref{fig:dnn_results_2} compares the performance vs. other semi-supervised graph-based schemes in the literature for sequential training.  Parallel results are shown in ~\ref{fig:mxn_val_acc} and~\ref{fig:mxn_val_wall_clock} for a 5\% labeled scenario. A more aggressive learning rate for a higher number of workers combined with data parallelism allows us to achieve faster accuracies for a given wallclock time.}
\label{}
\end{figure*}
For the experiments in this paper,  %\footnote{To test our model, we also initially experimented on a much smaller dataset consisting of Japanese vowels\cite{kudo1999multidimensional}, obtained from the UCI repository. Owing to the very small size, we do not report those results here}, 
  we use the TIMIT speech corpus~\cite{garofolo1993darpa} and report the frame-level phone classification accuracy. The training set consists of over $1$ million speech frames, each frame being a  $351$-d vector of cepstral coefficients. The output is mapped to a distribution over $39$ classes during scoring.  For both the sequential and parallel case, 
we experiment with  label ratios of $2\%,~5\%,~10\%,~30\%,~50\%$ and $100\%$ by randomly dropping labels from our training set.  For the $k$-NN graph construction, we set $k=10$ for all the experiments and use the Scikit machine learning library \cite{pedregosa2011scikit} that constructs the graphs using a fast ball-tree search. After symmetrization, affinities are computed by applying a radial basis function (RBF) kernel, such that  each entry $w_{ij}$ in the affinity matrix $W$, $w_{ij} = e^{-\frac{||x_i-x_j||}{2\sigma^2}}$. 
For graph partitioning, we use the  METIS graph partitioning library~\cite{karypis1998multilevelk} that uses a fast recursive multi-way partitioning algorithm to give approximately balanced blocks. Meta-batches are synthesized from the graph partitions as described in the previous sections; both partitioning and meta-batch synthesis are one-time pre-processing steps.
We originally implemented all our sequential models using the Theano toolkit~\cite{bergstra2010theano}  %which also automatically generates  code for GPU execution.
and for the parallel case, we implemented our algorithms in the  MXNet framework~\cite{mxnet} that provides parallel SGD functionality for both synchronous and asynchronous SGD (though we only use the synchronous version in our experiments).  For the results reported here we used the AdaGrad~\cite{duchi2011adaptive} variant of gradient descent. 
%The inherent parallel nature of our algorithm meant that we only had to write parallel data iterators for selecting mini-batches and the associated chunks of the affinity matrix at each worker, much like one would parallelize a fully supervised  learner.
We used a  DNN with four hidden layers, each 2000 units wide, using Rectified Linear Units~\cite{zeiler2013rectified} as the non-linear activation function, and a softmax output layer. We used dropout while training, reporting the results for the case when dropout probability is $0.2$. The model and hyperparameters are all kept unchanged  between parallel and  sequential versions.

Detailed results for the sequential training case are described in~\cite{sunil_mlslp_2016} and we reproduce one of the results to illustrate the efficacy of the method (see Figure~\ref{fig:dnn_results_2}). In the low labeled scenario, our method significantly outperforms  similar graph-based (sequential) SSL  methods described in~\cite{liu2013graph,malkin2009semi}.  For the parallel training case, we experimented with $2$, $4$ and $8$ workers, all running on GPU-enabled machines. We keep the batch size the same irrespective of the number of workers, implying that with more workers we average the gradients over a larger number of points. This obviously leads to fewer gradient updates per epoch, compared to the sequential case. However, since  gradients are less noisy when averaged over a larger number of training points,  in the parallel case we can be a little more aggressive with our learning rate. In our experiments we use a base learning rate of $0.001$ and an effective initial learning rate of $0.001k$ where $k$ is the number of parallel workers. After a fixed number of epochs (10 in our case), we reset the learning rate back to $0.001$.  Thus even the though sequential version has a larger number of gradient updates per epoch, as we can see in Figure~\ref{fig:mxn_val_acc}, the parallel training run is able to achieve higher validation accuracies per epoch due to a higher learning rate.  A more important metric for  parallel machine learning in general is the speed of learning -- in terms of validation loss -- as a function of wall-clock time. We show this in Figure~\ref{fig:mxn_val_wall_clock} where we see that the eight worker case is able to achieve significantly higher accuracies early on. We should note that synchronizing with the parameter server does introduce an overhead in our experiments -- in MXNet, for our implementation, we observe this to be a constant factor of two. That is an individual worker node in the parallel setting was about twice as slow compared to a worker in the sequential version, in terms of sample processing throughput. With a moderate number of workers, we are able to compensate for this overhead, and generally, the method scales well in the data-parallel paradigm, and we expect to see even larger speed ups as we increase the worker count.

\section{Conclusion}
We presented an efficient, scalable method for parallel graph-based SSL learning that can be applied in general to any parametric learner.  The methods presented were heuristically motivated; for our current research we are looking at further analysis, asynchronous versions of SGD and more provably optimal methods for constructing meta-batches. 

\bibliographystyle{plain}
\bibliography{deep_ssl}

\begin{thebibliography}{10}

\bibitem{bergstra2010theano}
James Bergstra, Olivier Breuleux, Fr{\'e}d{\'e}ric Bastien, Pascal Lamblin,
  Razvan Pascanu, Guillaume Desjardins, Joseph Turian, David Warde-Farley, and
  Yoshua Bengio.
\newblock Theano: a cpu and gpu math expression compiler.
\newblock In {\em Proceedings of the Python for Scientific Computing Conference
  (SciPy)}, volume~4, page~3. Austin, TX, 2010.

\bibitem{chapelle2006semi}
Olivier Chapelle, Bernhard Sch{\"o}lkopf, Alexander Zien, et~al.
\newblock {\em Semi-supervised learning}.
\newblock MIT press Cambridge, 2006.

\bibitem{mxnet}
Tianqi Chen, Mu~Li, Yutian Li, Min Lin, Naiyan Wang, Minjie Wang, Tianjun Xiao,
  Bing Xu, Chiyuan Zhang, and Zheng Zhang.
\newblock Mxnet: {A} flexible and efficient machine learning library for
  heterogeneous distributed systems.
\newblock {\em CoRR}, abs/1512.01274, 2015.

\bibitem{duchi2011adaptive}
John Duchi, Elad Hazan, and Yoram Singer.
\newblock Adaptive subgradient methods for online learning and stochastic
  optimization.
\newblock {\em The Journal of Machine Learning Research}, 12:2121--2159, 2011.

\bibitem{garofolo1993darpa}
John~S Garofolo, Lori~F Lamel, William~M Fisher, Jonathan~G Fiscus, and David~S
  Pallett.
\newblock Darpa timit acoustic-phonetic continous speech corpus cd-rom. nist
  speech disc 1-1.1.
\newblock {\em NASA STI/Recon Technical Report N}, 93, 1993.

\bibitem{kaiwei2015nipsparallelworkshop}
{K. Wei, R. Iyer, S. Wang, W. Bai, J. Bilmes}.
\newblock How to intelligently distribute training data to multiple compute
  nodes: Distributed machine learning via submodular partitioning.
\newblock In {\em Neural Information Processing Society (NIPS) Workshop},
  Montreal, Canada, December 2015.
\newblock LearningSys Workshop, \url{http://learningsys.org}.

\bibitem{karypis1998multilevelk}
George Karypis and Vipin Kumar.
\newblock Multilevel k-way partitioning scheme for irregular graphs.
\newblock {\em Journal of Parallel and Distributed computing}, 48(1):96--129,
  1998.

\bibitem{smola2015}
Mu~Li, Dave~G. Andersen, and Alexander~J. Smola.
\newblock Graph partitioning via parallel submodular approximation to
  accelerate distributed machine learning.
\newblock {\em CoRR}, abs/1505.04636, 2015.

\bibitem{liu2013graph}
Yuzong Liu and Katrin Kirchhoff.
\newblock Graph-based semi-supervised learning for phone and segment
  classification.
\newblock In {\em INTERSPEECH}, pages 1840--1843, 2013.

\bibitem{malkin2009semi}
Jonathan Malkin, Amarnag Subramanya, and Jeff~A Bilmes.
\newblock On the semi-supervised learning of multi-layered perceptrons.
\newblock In {\em INTERSPEECH}, pages 660--663, 2009.

\bibitem{pedregosa2011scikit}
Fabian Pedregosa, Ga{\"e}l Varoquaux, Alexandre Gramfort, Vincent Michel,
  Bertrand Thirion, Olivier Grisel, Mathieu Blondel, Peter Prettenhofer, Ron
  Weiss, Vincent Dubourg, et~al.
\newblock Scikit-learn: Machine learning in python.
\newblock {\em The Journal of Machine Learning Research}, 12:2825--2830, 2011.

\bibitem{subramanya2009entropic}
Amarnag Subramanya and Jeff~A Bilmes.
\newblock Entropic graph regularization in non-parametric semi-supervised
  classification.
\newblock In {\em Advances in Neural Information Processing Systems}, pages
  1803--1811, 2009.

\bibitem{sunil_mlslp_2016}
Sunil Thulasidasan and Jeff Bilmes.
\newblock Semi-supervised phone classification using deep neural networks and
  stochastic graph-based entropic regularization.
\newblock In {\em 2016 Workshop on Machine Learning in Speech and Language
  Processing}, San Francisco, CA, September 2016.

\bibitem{zeiler2013rectified}
Matthew~D Zeiler, Marc'Aurelio Ranzato, Rajat Monga, Min Mao, Kun Yang,
  Quoc~Viet Le, Patrick Nguyen, Alan Senior, Vincent Vanhoucke, Jeffrey Dean,
  et~al.
\newblock On rectified linear units for speech processing.
\newblock In {\em Acoustics, Speech and Signal Processing (ICASSP), 2013 IEEE
  International Conference on}, pages 3517--3521. IEEE, 2013.

\bibitem{zhu2005semi}
Xiaojin Zhu, John Lafferty, and Ronald Rosenfeld.
\newblock {\em Semi-supervised learning with graphs}.
\newblock Carnegie Mellon University, Language Technologies Institute, School
  of Computer Science, 2005.

\end{thebibliography}

\end{document}